\documentclass{article}
\usepackage{spconf,amsmath,graphicx}
\usepackage{qtree,booktabs}
\usepackage{xcolor}
\usepackage{framed}
\usepackage{hyperref}

\newcommand{\abdo}[1]{}

\def\x{{\mathbf x}}

\title{STOP: A dataset for Spoken Task Oriented Semantic Parsing}
\name{%
\begin{tabular}{c}
Paden Tomasello,
Akshat Shrivastava,
Daniel Lazar,
Po-Chun Hsu,
Duc Le,
Adithya Sagar, \\
Ali Elkahky,
Jade Copet,
Wei-Ning Hsu,
Yossi Adi,
Robin Algayres,
Tu Ahn Nguyen, \\
Emmanuel Dupoux,
Luke Zettlemoyer,
Abdelrahman Mohamed
\end{tabular}}
\address{Meta AI}
\copyrightnotice{978-1-6654-7189-3/22/\$31.00~\copyright2023 IEEE}
\begin{document}

%
%
%

\maketitle

\begin{abstract}


End-to-end spoken language understanding (SLU) predicts intent directly from audio using a single model. It promises to improve the performance of assistant systems by leveraging acoustic information lost in the intermediate textual representation and preventing cascading errors from Automatic Speech Recognition (ASR). Further, having one unified model has efficiency advantages when deploying assistant systems on-device. However, the limited number of public audio datasets with semantic parse labels hinders the research progress in this area. In this paper, we release the Spoken Task-Oriented semantic Parsing (STOP) dataset \footnote{The dataset can be downloaded at \url{https://github.com/facebookresearch/fairseq/tree/main/examples/audio_nlp/nlu}}, the largest and most complex SLU dataset publicly available.
Additionally, we define low-resource splits to establish a benchmark for improving SLU when limited labeled data is available. Furthermore, in addition to the human-recorded audio, we are releasing a TTS-generated versions to benchmark the performance for low-resource and domain adaptation of end-to-end SLU systems.

\end{abstract}


\begin{figure}[t]
\begin{framed}
\caption{Semantic Parse from TOPv2 }
\label{fig:semantic_parse}
\textbf{Utterance:} Directions to the Eagles game \\
\textbf{Semantic Parse:} [IN:GET\_DIRECTIONS Directions to
[SL:DESTINATION [IN:GET\_EVENT the [SL:NAME\_EVENT
Eagles ] [SL:CAT\_EVENT game ] ] ] ] \\
\textbf{Tree Representation:} \\ \\
\Tree[.IN:GET\_DIRECTIONS [.\textit{ Directions to } ]
          [.SL:DESTINATION [.IN:GET\_EVENT
                                [.\textit{the} [.SL:NAME\_EVENT \textit{Eagles} ] [.SL\_CAT\_EVENT \textit{game} ]]
                            ]
        ]]
\end{framed}
\end{figure}

\section{Introduction}

Assistant systems are becoming integrated into an increasing number of out daily devices, from cell phones and smart speakers to cars and televisions. Traditional assistant systems utilize a cascaded approach: an Automatic Speech Recognition (ASR) system transcribes the audio to text, followed by a separate Natural Language Understanding (NLU) system which predicts the user's intentions from the decoded text. Although cascade systems enable using audio-only and text-only training resources, they are prone to error propagation when the ASR systems fail to generate the correct transcript. Further, an NLU model fed with decoded text cannot leverage paralinguistic information in the input, e.g., pausing, intonation, and stress variations. Moreover, two separate models in a cascade system increase the computational requirements for on-device applications, as well as duplicate efforts in building and maintaining them. 

End-to-end spoken language understanding (SLU) has recently shown great promise by directly predicting intent and slots values from audio using a single system \cite{serdyuk} \cite{haghani}. It resolves the ambiguity of similarly sounding inputs that would be competing hypotheses of an ASR system. SLU has the advantage of leveraging acoustic information lost in the cascaded approach, like prosody and rhythm. One of the challenges for SLU research today is a lack of large and challenging audio datasets with semantic parse labels. In this paper, we release the Spoken Task-Oriented Semantic Parsing (STOP) dataset, which contains 1) three times as many audio samples as existing datasets 2) five times as many speakers as existing datasets 3) and compositional queries to support the ever-increasing demands of supporting real-world assistant systems. 


We recognize that collecting audio and intent pairs is a challenging and resource intensive task. Speech audio could contain sensitive information and proper precaution needs to be taken during collection. These privacy considerations pose a challenge to large-scale dataset collection. Accordingly, we aim to establish the first-ever low-resource splits for an end-to-end SLU dataset to support research on training with limited training data. Our baselines show the importance of audio and ASR pretraining for obtaining competitive performance versus cascaded systems. 

\abdo{stress on the experimental contribution of this paper for out-of-domain genralization using self-supervised models}

\abdo{intro needs some discussion on the the domain expansion problem and how this benchmark address that using TTS data, which is common practice but not evaluated in a publicly available benchmark before}

\abdo{say something about the models that we are releasing with the data for reproducing our results}

We summarize our contributions as follows:
\begin{itemize}
    \item Release of the STOP dataset, the largest and most semantically complex end-to-end spoken language understanding dataset publicly available,
    \item Defining low-resource splits as a benchmark for training with limited natural speech data samples,
    \item TTS-generated versions of the dataset for benchmarking low-resource training
    \item Experimental analysis on training end-to-end spoken language understanding models, with insights into speech pretraining, performance in the low-resource setting, domain adaptation, and how TTS audio can be incorporated into these settings.
    \item Open source release of our models and code for reproduction at \url{https://github.com/facebookresearch/fairseq/tree/main/examples/audio_nlp/nlu}
\end{itemize}

We conclude with promising future directions of research.



\section{Related work}
In the 1990s, the first audio dataset containing semantic labels was first released in the Air Travel Information System (ATIS) corpus \cite{hemphill}, followed by the Switchboard-DAMSL Labeling Project \cite{jurafsky}. 
The successful application of sequence-to-sequence neural models to machine translation systems \cite{sutskever} and speech recognition \cite{chan} ignited the interest in end-to-end systems for SLU, which were introduced in \cite{serdyuk} and \cite{haghani}. Due to the scarcity of audio to intent data for the ever-increasing number of domains in practical assistant scenarios, transfer learning has been an area of active research explored in \cite{schuster}, \cite{tomashenko} as well as ASR Pretraining in \cite{lugosch}.

More recently, there have been several efforts around newer, larger datasets for SLU, namely, the Snips benchmark \cite{coucke}, Fluent Speech Command (FSC) corpus \cite{lugosch}, and the SLURP dataset \cite{bastianelli}. In this paper, we introduce the STOP dataset, which contains three times as many audio files and five times as many speakers compared to the largest of these datasets, SLURP. Further, our dataset includes compositional queries with nested intents, which no previous publicly available SLU dataset included. Compositional or nested queries for text-based systems were first introduced in \cite{gupta} with the first version of the TOP dataset with 2 domains. TOPv2 was release in \cite{chen}, which adds 6 additional domains for a total of 137k more samples across 8 domains.

\section{Datasets}

\subsection{STOP Data Collection and Verification} 

The STOP dataset builds upon the TOPv2 dataset \cite{chen} which provides text-only inputs and target semantic parse trees. For every utterance-semantic parse pair, we collect an audio recording of the input utterance by requesting workers through Amazon's Mechanical Turk to record themselves. We ran the recordings through an ASR system to control the quality of the recorded audio. In our case, it is the fine-tuned wav2vec 2.0 \cite{baevksi} ASR model in the HuggingFace \cite{wolf} library. Recordings are automatically added to the dataset only if their character error rate is below 50\% from the expected input text.
Given that the wav2vec 2.0 model is trained on English audiobooks data, it performed poorly for non-native speakers, so any sample below the threshold was reviewed manually. In the event that the recording does not match the utterance it is discarded and recollected by another Mechanical Turk worker.


\subsection{Normalizing text to spoken form}
\label{section:normalization}

The TOPv2 dataset contains utterances in written text form, which may include digits or abbreviations like ``4" and ``CA." However, ASR systems are typically trained on spoken text forms like ``four" and ``California," respectively. To make our dataset more aligned with ASR system, we used an in-house spoken text normalization tool to enumerate possible pronunciations for each utterance and an in-house ASR model to predict the audio transcription. We then computed the edit distances across the possible normalizations and chose the normalization with the lowest distance. We include both the normalized and unnormalized form for each sample.

\subsection{Low resource splits}
\label{section:low-resource-spits}
Collecting audio recordings and annotating them with their semantic parses is laborious. We, therefore, believe a crucial research area for end-to-end NLU systems is low resource training, where the amount of training data is limited. Therefore, for our STOP dataset, in addition to the complete training set, we are also releasing low resource splits, which can be used as baselines for low-resource end-to-end NLU training and target datasets for domain adaptation. 
TOPv2 \cite{chen} defined \textbf{SPIS}, or \textit{samples per intent and slot}, which indicates the number of training samples available for each slot and intent and establishes a benchmark for training with only 25 SPIS.



\subsection{TTS TOPv2} \label{sec-tts-topv2} 

Going deeper in simulating real-world pre-release conditions in the STOP benchmark, we are releasing two different TTS versions of TOPv2. First, the utterances in TOPv2 are converted to spoken form as detailed in Section \ref{section:normalization}. We then run the VITS \cite{jaehyeon} text-to-speech (TTS) system trained on LibriTTS \cite{heiga} and the CSTR VCTK Corpus \cite{yamagishi2019vctk}. In the low-resource and domain adaptation parts of Section \ref{experiments}, the performance of NLU models trained with these machine-generated datasets is compared to ones trained on natural human-spoken audio. Although it is a common practice in the industry to use TTS-generated speech for enabling pre-release NLU domains, STOP is the first publicly-available benchmark to offer the opportunity to academic institutions to conduct research in this area.


\subsection{Dataset statistics and comparisons} 

We collected two audio recordings for every utterance in the validation and test sets of TOPv2 and one for each of the utterances in the train set. Speakers may only be in one split to ensure no data pollution exists between different partitions. Additionally, we limit the number of recordings any given speaker can submit to 3000. 
We ask speakers to self-identify their gender with options, Male, Female, or prefer not to say, or non-binary. Table \ref{tab:gender_stats} shows dataset statistics across different genders. Also, we ask speakers to self-identify whether they are English-native speakers, non-native speakers, or prefer not to say. The distribution can be seen in Table \ref{tab:native_english_stats}




\begin{table}[h]
  \caption{ Gender Statistics  }
  \label{tab:gender_stats}
  \centering
  \begin{tabular}{cccc}
    \toprule
    \textbf{Female} & \textbf{Male} & \textbf{Non-binary} & \textbf{Unanswered} \\
    \midrule
    44.55\% & 54.62\% &  0.79\%  & 0.04\% \\
    \bottomrule
  \end{tabular}
\end{table}

\begin{table}[h]
  \caption{ Native English Statistics }
  \label{tab:native_english_stats}
  \centering
  \begin{tabular}{ccc}
    \toprule
    \textbf{Native} & \textbf{Non-native} & \textbf{Prefer not to say}  \\
    \midrule
    95.18\% & 4.09\% & 0.73\% \\
    \bottomrule
  \end{tabular}
\end{table}

\begin{table}[h]
  \caption{ Spoken dataset comparison }
  \label{tab:dataset_comparison}
  \centering
  \begin{tabular}{lcccc}
    \toprule
    & FSC & SNIPS & SLURP & \textbf{STOP} \\
    \midrule
    Speakers & 97 & 67 & 177 & \textbf{885} \\
    Audio Files & 30043 & 5886 & 72277 & \textbf{236477} \\
    Duration[hrs] & 19 & 5.5 & 58 & \textbf{218} \\
    \bottomrule
  \end{tabular}
\end{table}

Overall, our STOP dataset is three times as large as previous datasets and contains five times as many speakers. A complete comparison with other Spoken NLU datasets is in Table \ref{tab:dataset_comparison}. 

\section{Experiments} \label{experiments}
\subsection{Metrics}

We use two metrics when evaluating our models. The first is the Exact Match (EM) accuracy, which reports the model's accuracy in perfectly reproducing the semantic parse tree with the correct intent and slot tags. The second is the Exact Match Tree (EM-Tree), which reports the accuracy of producing the parse trees' intent and slot labels but does not evaluate the tags' correctness. 
In End-to-End NLU, EM-Tree is a good proxy if the model understands the intent and the structure of the user interaction but cannot transcribe the audio perfectly, e.g., ASR errors in rare entity names. In contrast, EM represents understanding the intent and being able to transcribe the audio perfectly. 


\subsection{Models}

Our experiments contain baseline results using cascaded and end-to-end NLU systems. The cascaded system first transcribes the audio using an ASR system, which is a pre-trained wav2vec2.0 model fine-tuned on either 960h of Librispeech \cite{panayotov} or the full STOP training set. The text is then fed into an NLU model that is a pretrained BART-base model \cite{bart} containing a 6 layer transformer encoder and 6 layer transformer decoder, augmented with a copy-generate mechanism \cite{decoupled}.

The end-to-end baseline system is an encoder-decoder Transformer architecture with an output vocabulary containing letters and all the intents and parses in STOP. For training, we use an Adam optimizer \cite{kingma} with the betas set to 0.9 and 0.98. We use a tri-stage learning rate scheduler \cite{park}, where the learning rate is warmed up for the first
10\% of updates to a maximum of 1e-4, held constant for the next 40\% and then linearly decayed for the remainder updates to 5e-6. We trained all our models with a maximum number of updates of 320,000 and chose the best checkpoint based on the validation set.



\subsection{In-domain vs Out-of-domain ASR}

Given the surge of publicly-available ASR models trained on public data, we sought to answer how well these models perform for the assistant domain. Including out-of-domain experiments in our benchmark opens the space for research work in transfer learning for end-to-end SLU models. We used publicly-available Wav2Vec2.0\cite{baevksi} and HuBERT\cite{hsu} pre-trained models, fine-tuned using a CTC loss\cite{graves} on Librispeech 960h and evaluated them on the STOP eval and test set in addition to the Libripeech test and dev sets. We then took the same models and fine-tuned them using the STOP training set to perform the same evaluation.


\subsection{Speech pre-training for NLU}

Similar to \cite{lugosch}, we demonstrate that speech pre-training is a critical component for training end-to-end NLU systems. This work evaluates two speech pre-training methods, namely Wav2Vec 2.0 \cite{baevksi} and HuBERT \cite{hsu}. 

\subsection{ASR pre-training}

In cascaded NLU systems, the semantic parse model can benefit significantly from employing a pre-training like BART \cite{bart}. Currently audio pre-training is encoder only, leaving the decoder randomly initialized when starting the training for a downstream task in a seq2seq manner. This work also benchmarks the benefit of supervised pre-training of the decoder component. As a secondary pre-training task, we train a seq2seq ASR model that uses the pre-trained encoder. We present systems with ASR pre-training utilizing the STOP and the larger Librispeech datasets. 


\subsection{Low resource end-to-end NLU}
\label{section:experiments-low-resource}

Since collecting audio samples with the corresponding parse labels can be impractical due to a variety of concerns, 
we investigate how much audio intent pairs are needed to build an end-to-end system for a single domain. To this end, we train a HuBERT encoder on the Librispeech training set, then fine-tune it on the held-in domain (either Weather or Reminder). We leverage the low resource splits in \ref{section:low-resource-spits}, with SPIS in$\{10, 25, 50, 100, 250, 500, 1000\}$, and analyze the error rates with progressively more data.

\subsection{Domain adaptation for end-to-end NLU models}
\label{section:experiments-domain-adaptation}

One way of improving the performance of models trained in the low-resource setting is to perform source training on held in domains, then fine-tune on the low-resource target domain. Similar to the original TOPv2 paper, we study how end-to-end NLU models adapt to new domains.

We study this method by first taking the HuBERT encoder trained on Librispeech then source training on all data samples from ``held-in'' TOPv2 domains -- all domains aside from Weather and Reminder as described in Section \ref{section:low-resource-spits}. We then fine-tune our model using the low-resource splits with varying amounts of data for both the Weather and the Reminder domain. We compare our results to the results of \ref{section:experiments-low-resource} to see how our models benefits from having previously been train on semantic parse data from different domains.

\subsection{Leveraging TTS for zero resource and low resource end-to-end NLU}
\label{section:experiments-tts-low-resource}

Another way to overcome the cost of collecting large natural datasets is to supplement or replace the training data with synthetic high resource data, which is much cheaper to generate than natural speech. Accordingly, we explore the feasiblity of building end-to-end NLU systems using only TTS data, or leveraging mostly TTS with a small amount of audio and intent pairs from our low-resource splits. To do so, we use the TTS versions of TOPv2 details in Section \ref{sec-tts-topv2}, optionally combined with the low-resource training split detailed in section \ref{section:low-resource-spits}. In this analysis, all data is from the held-out domain considered (Weather or Reminder).



A comparison of our these ablation studies can be seen in Table \ref{tab:tts-low-resource-domain-adaptation-experiment-comparison}

\begin{table}[ht]
    \centering
    \begin{tabular}{l|cc}
        \toprule
         \textbf{Experiment} & \textbf{Parse Data?} & \textbf{TTS Data?} \\
        \midrule
         Low-resource Baseline & No & No \\
         Domain Adaptation & Yes & No  \\
         TTS - Low-resource & No & Yes 
    \end{tabular}
    \caption{\small A comparison of low-resource and domain adaptation ablation studies for the Weather and Reminder domains. }
    \label{tab:tts-low-resource-domain-adaptation-experiment-comparison}
\end{table}

\section{Results and Analysis}

\begin{table*}[ht]
  \caption{ ASR Results }
  \label{tab:asr_results}
  \centering
  \begin{tabular}{lccccccc}
    \toprule
    \textbf{Encoder} & \textbf{Training Set} & \textbf{dev\_other} & \textbf{dev\_clean}  & \textbf{test\_clean} & \textbf{test\_other} & \textbf{stop\_eval} & \textbf{stop\_test} \\
    \midrule
    HuBERT & Librispeech & \textbf{8.47} & \textbf{2.99} & \textbf{3.25} & \textbf{8.06} & 25.68 & 26.19 \\
    Wav2Vec2.0 & Librispeech & 9.02 & 3.31 & 3.42 & 8.81 & 26.53 & 26.93\\
    \midrule
    HuBERT & STOP & 46.31 & 31.30 & 31.52 & 47.16 & \textbf{4.29} & \textbf{4.26}\\
    Wav2Vec2.0 & STOP & 48.67 & 32.51 & 33.04 & 49.71 & 4.43 & 4.45 \\
    \bottomrule
  \end{tabular}
\end{table*}

\begin{table}[h]
  \caption{ Speech Pre-training }
  \label{tab:speech_pretraining}
  \centering
  \begin{tabular}{lcc}
    \toprule
    \textbf{Encoder}  & \textbf{Valid (EM/Tree)}  & \textbf{Test (EM/Tree)} \\
    \midrule
    End-to-end \\
    \midrule
    Random  & 35.28 / 57.73 & 36.54 / 57.01 \\
    Wav2Vec2.0 & 67.71 / 83.35 & 68.05 / 82.53\\
    HuBERT & 67.86 / 83.43 & 68.40 / 82.85\\
    \midrule
    \midrule
    Cascaded \\
    \midrule
    STOP ASR & 72.43 / 86.58 & 72.36 / 85.77 \\
    Librispeech ASR& 24.14 / 49.32 & 31.32 / 60.70 \\
    Text GT & 85.65 / 88.43 & 85.25 / 87.85 \\
    \bottomrule
  \end{tabular}
\end{table}

\subsection{ ASR Results for In-domain vs Out-of-domain data }

Table \ref{tab:asr_results} shows the results of training ASR models on Librispeech, STOP and a combined dataset. As can be seen, training purely on LibriSpeech performs quite poorly on the STOP eval and test sets, with WER above 25\%. By fine-tuning on the STOP training set however, we are able to achieve the best WERs on the STOP eval and test sets of 4.29 and 4.26 respectively using a HuBERT encoder model. We hope this result demonstrates the significance of our dataset outside of end-to-end systems assistant systems, since it is the largest in-domain assistant ASR dataset as far as we know.

\subsection{Speech pre-training for NLU}

As can be seen in Table \ref{tab:speech_pretraining}, speech pre-training is a critical component for end-to-end NLU. Compared to using a randomly initialized model, Wav2Vec2.0 and HuBERT have almost twice the EM accuracy for both valid and test sets. For Tree accuracy, Wav2Vec2.0 and HuBERT improve upon the randomly initialized model by roughly 25\% points. 


\begin{table}[t]
  \caption{ ASR Pretraining}
  \label{tab:asr_pretraining}
  \centering
  \begin{tabular}{lcccc}
    \toprule
    & \textbf{Valid (EM/Tree)}  & \textbf{Test EM/Tree)}\\
    \midrule
    HuBERT \\
    \midrule 
    None & 67.86 / 83.43 & 68.40 / 82.85 \\
    Librispeech & 68.83 / 83.45 & \textbf{69.36} / 82.84 \\
    STOP & \textbf{69.12} / \textbf{83.89} & 69.23 / \textbf{82.87} \\
    \midrule
    \midrule
    wav2vec2.0 \\
    \midrule
    None & 67.71 / 83.35 & 68.05 / 82.53\\
    Librispeech & 68.24/83.55 & 68.47/82.49 \\
    STOP & 68.52 / 83.83 & 68.70 / 82.78 \\
    \midrule
    \midrule
    Cascaded \\
    \midrule
    STOP ASR & 72.43 / 86.58 & 72.36 / 85.77 \\
    LibriSpeech ASR & 24.14 / 49.32 & 31.32 / 60.70 \\
    GT Text & 85.65 / 88.43 & 85.25 / 87.85 \\
    \bottomrule
  \end{tabular}
\end{table}

\subsection{ ASR pre-training }

In Table \ref{tab:asr_pretraining}, we compare the results of performing ASR pretraining across Librspeech and STOP for both HuBERT and Wav2Vec2.0 encoders. As can be seen, leveraging ASR pre-training improves in all scenarios by about 1\%. Looking at HuBERT, the difference between Librispeech ASR pre-training and STOP ASR pretraining is small, with Librispeech performing slightly better on Test EM accuracy and STOP performing slightly better on Test Tree Accuracy. Similarly for Wav2Vec2.0, we see STOP ASR pre-training perform only slightly better than Librispeech pre-training for both Test EM accuracy and Tree Accuracy. We conclude that for decoder initialization, having some intelligent initialization obtained from ASR pre-training is adequate and does not require in-domain pre-training for decent performance gains.

\subsection{ Low resource Baseline }

\begin{figure}[h]
\includegraphics[width=\linewidth]{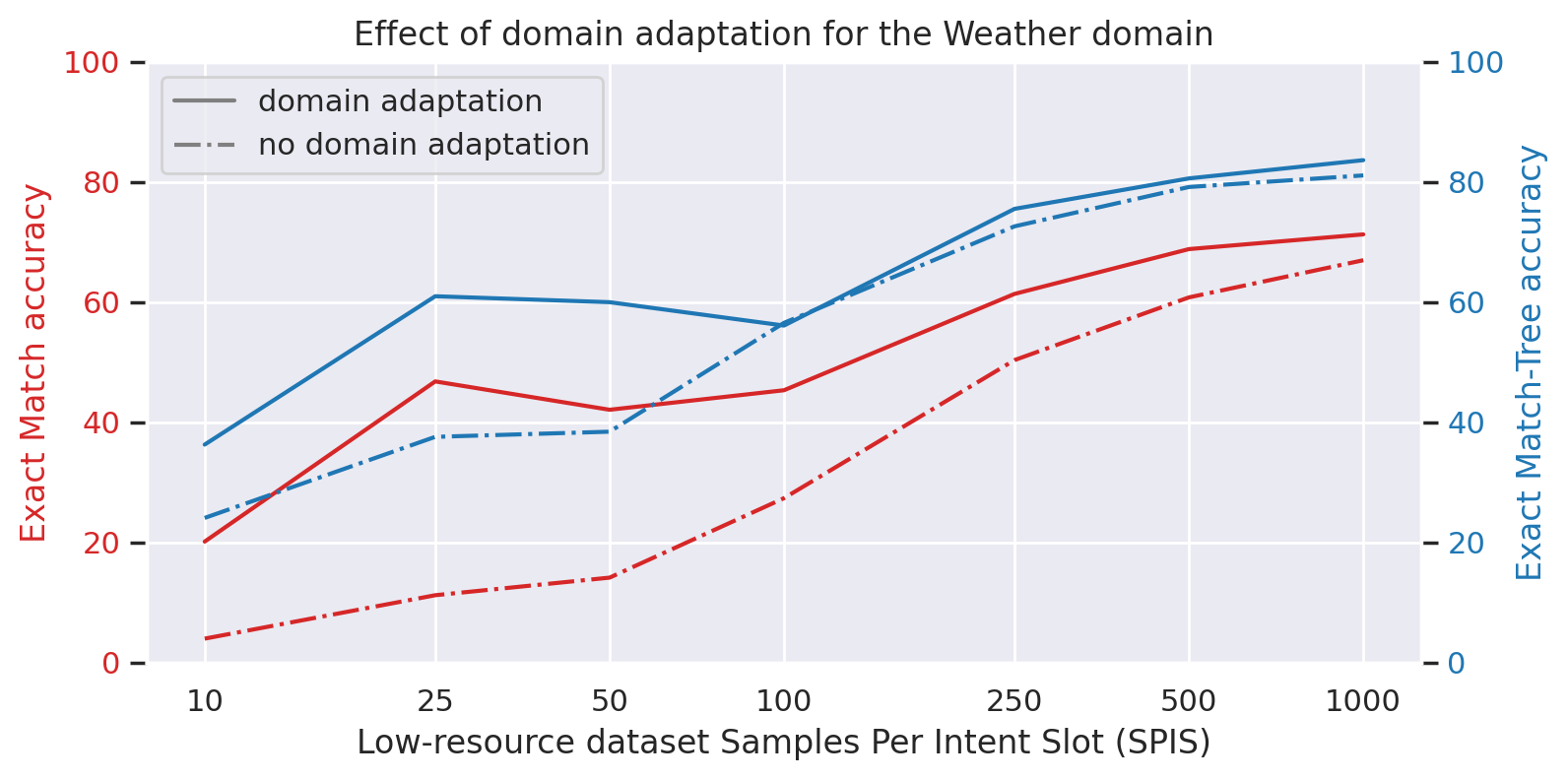}
\includegraphics[width=\linewidth]{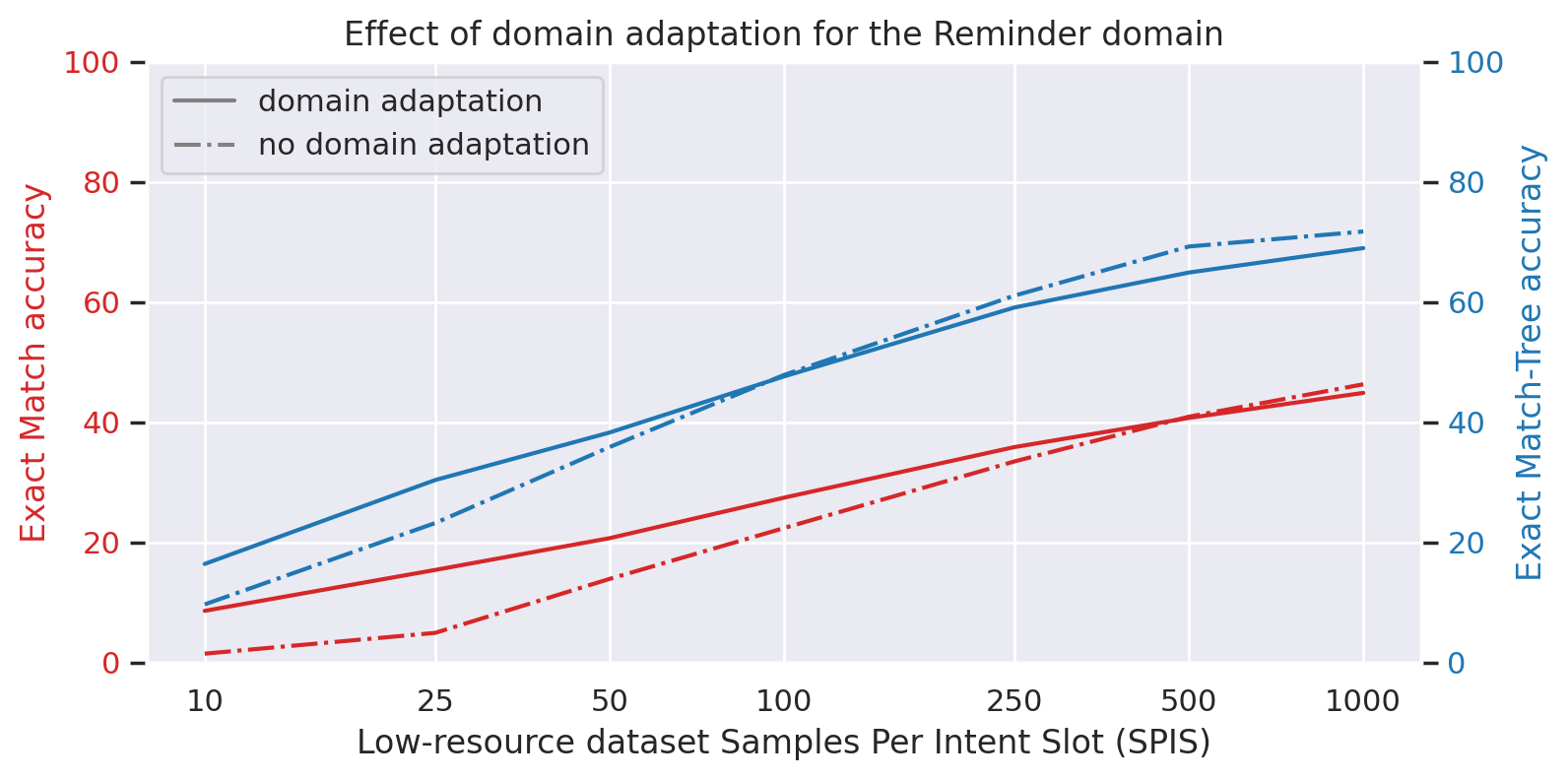}
\caption{\small Low resource training in the Weather and Reminder domains, compared to low-resource training after source training on high resource held-in domains (domain adaptation). Source training generally improves downstream target performance. }
\label{fig:domain_adaptation}
\end{figure}

As shown in Figure \ref{fig:domain_adaptation}, we find that scaling up training data greatly improves training model accuracy. EM and EM-tree trend together, though in Reminder, EM benefits more from scaling up training data than EM-tree. This may be a result of the compositional nature of the Reminder domain, as opposed to Weather, which has a more flat semantic structures.

\subsection{ Domain Adaptation }
We find that source training significantly increases downstream performance on the target task, especially for the very low resource regime (SPIS $<100$), and especially when Weather is the target domain. Interestingly, in the higher resource setting for Reminder, sometimes not performing source training performs better than source training. We speculate that some domains transfer well, while others interfere with downstream performance. Further study is needed to understand and mitigate this effect.

\subsection{TTS-Augmented Low Resource NLU }
\begin{figure}
\includegraphics[width=\linewidth]{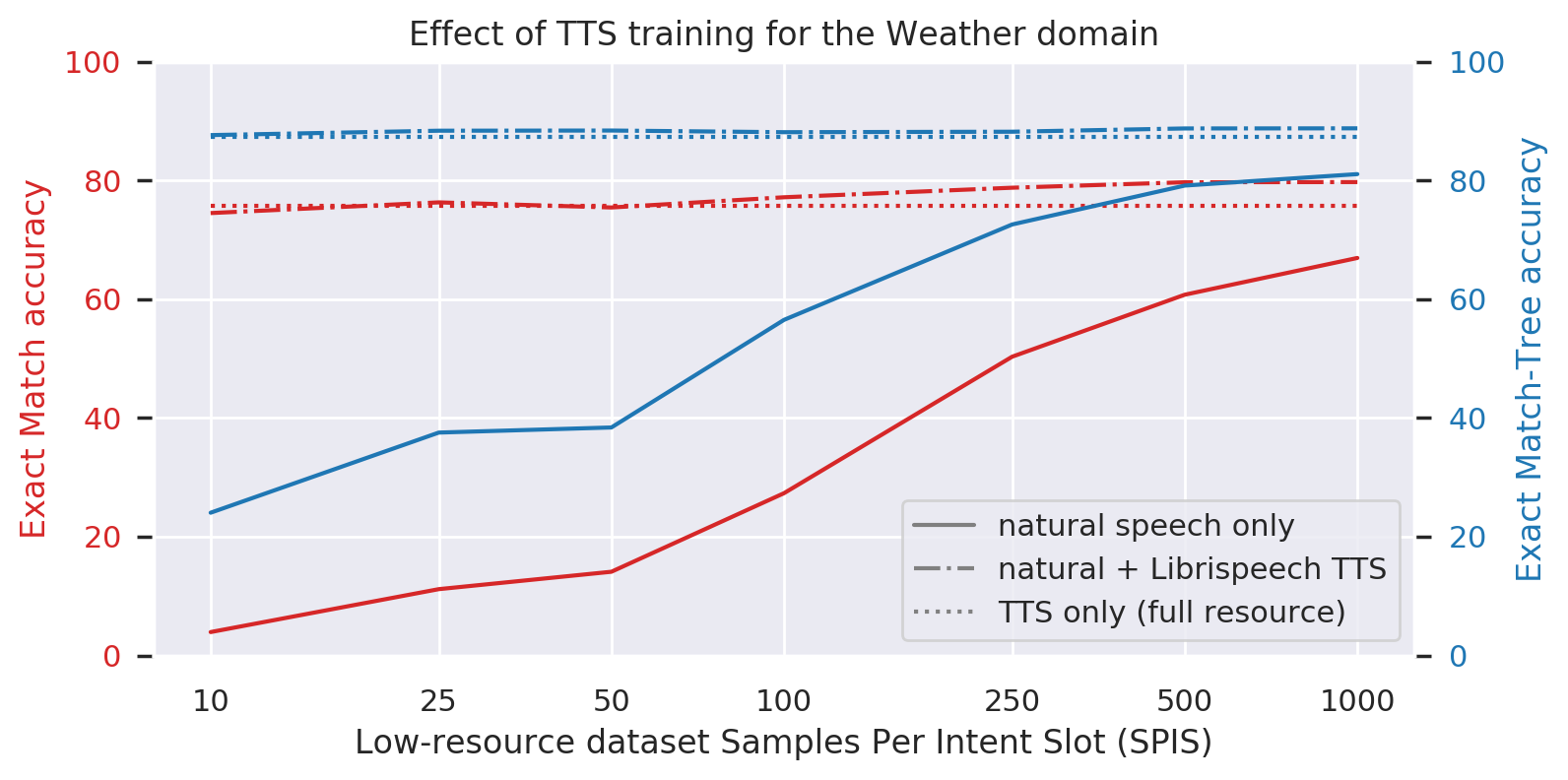}
\includegraphics[width=\linewidth]{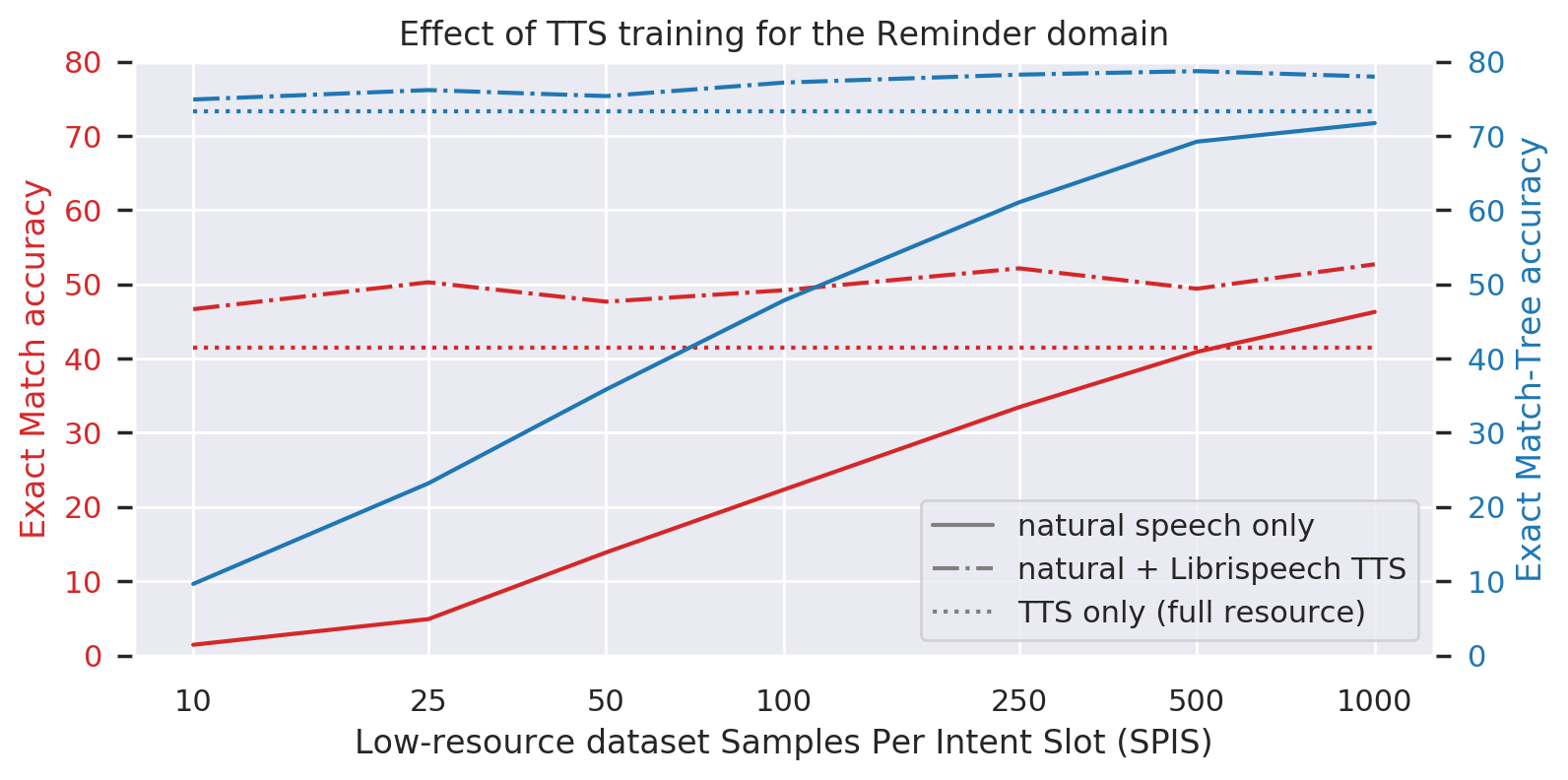}
\caption{\small Comparison of low-resource training with only low-resource natural speech, low-resource natural speech and full-resource TTS, and full-resource TTS only. TTS generally greatly improves model accuracy, and adding natural speech improves it further in almost all cases.}
\label{fig:low_resource_vs_tts}
\end{figure}

We compare three models: (1) trained on low-resource natural speech only, (2) trained on a combination of low-resource natural speech and high-resource synthetic speech data, and (3) trained only on high-resource synthetic speech. We find that even with no natural speech included, (3) performs better than (1). Adding natural speech further improves performance, and (2) always performs better than (1).

We again observe a difference between the Weather and Reminder domains: the gap between (1) and (3) is larger in Weather than Reminder, whereas adding natural speech to TTS improves Reminder more than Weather.

\begin{figure}
\includegraphics[width=\linewidth]{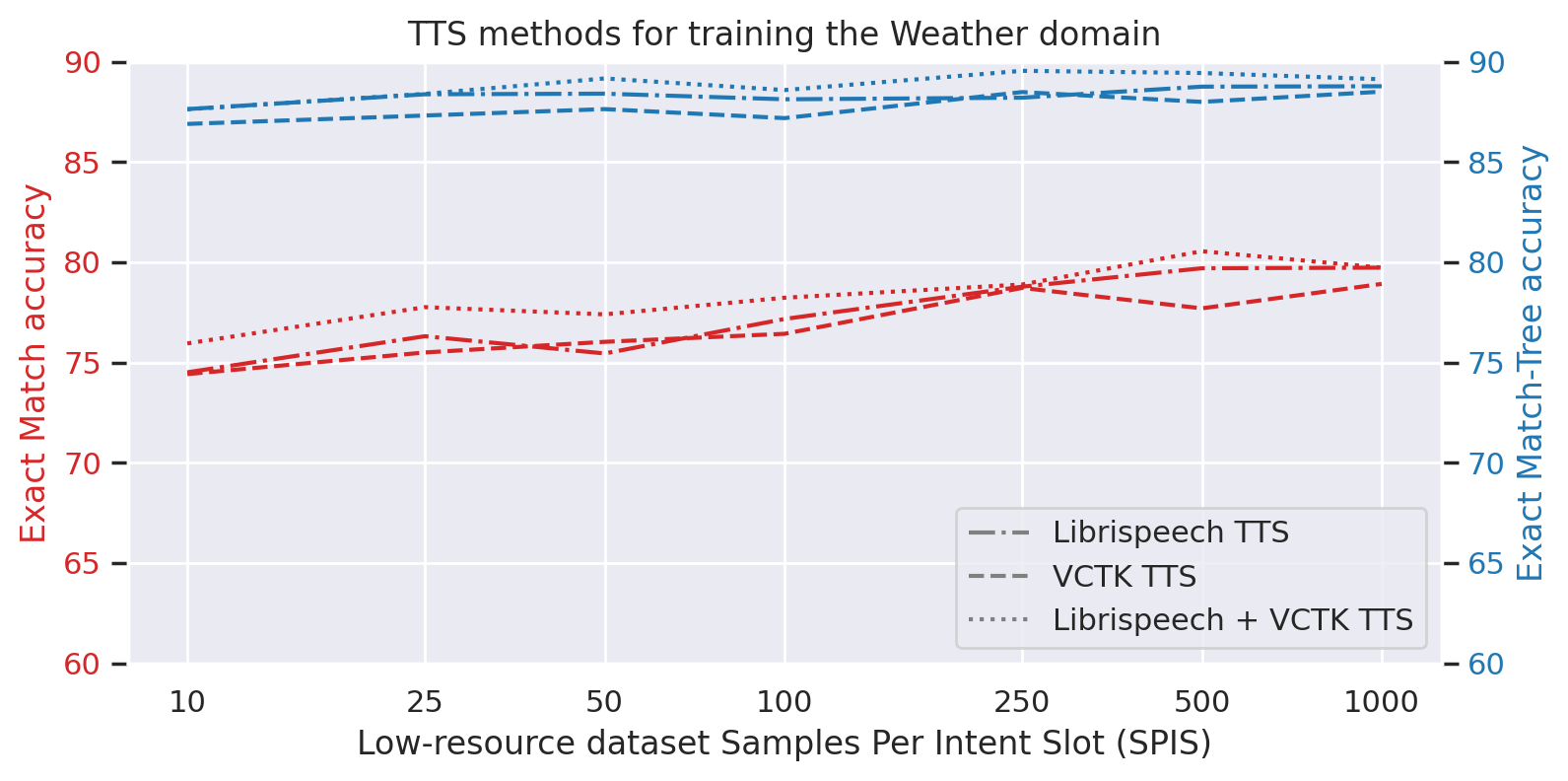}
\includegraphics[width=\linewidth]{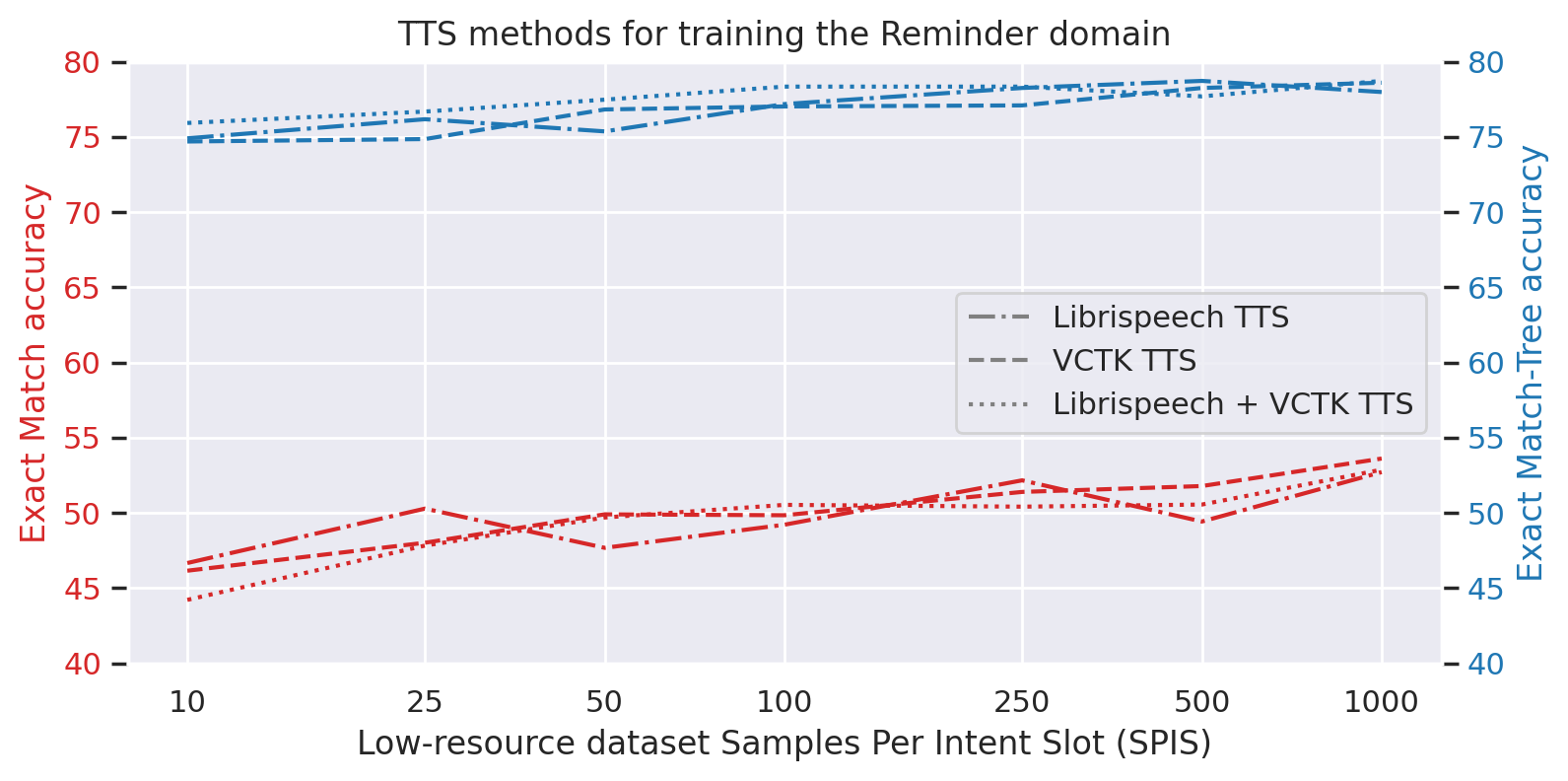}
\caption{\small Comparison of TTS methods for aiding low-resource training.}
\label{fig:tts_comparison}
\end{figure}

We also compare the performance of the two different TTS methods described in Section \ref{section:low-resource-spits} -- for convenience abbreviated as Librispeech and VCTK. We compare models trained on low-resource natural speech supplemented with high-resource Librispeech, VCTK, and the union of the two TTS datasets. We find that including both Librispeech and VCTK generally improves model performance beyond models trained on either of the two individually. This suggests that adding diversity to TTS data transfers to improved performance on natural speech test samples.

\section{Conclusion}

In this work, we: release STOP, the largest SLU dataset for end-to-end semantic parsing, provide initial baselines using both an end-to-end as well as a cascaded system, and show the importance of in-domain ASR training for cascaded systems. Additionally, we find ASR pre-training on top of audio pretraining can lead to improvements in E2E systems, closing the gap to cascaded systems. We believe there are a three particularly promising directions to explore so we may one day reach parity and eventually out-perform cascaded systems. First, is pseudolabeling or model distillation, where we can use a cascaded model to train a end-to-end model on large amounts of unlabeled audio. Second, is building more semantically meaningful representations. We hypothesize cascaded systems surpass E2E systems due to large text pre-training datasets. Third, is building a larger TTS dataset on-top of synthetic utterance-parse pairs. Our current results show impressive performance using only TTS data, and our current TTS data is built only on-top of the TOPv2 utterances, but could be expanded further easily. Finally, while this work did not explore on-device models, we believe end-to-end NLU is especially promising in compute and memory limited environments.



\bibliographystyle{IEEEtran}

\bibliography{refs}


\end{document}